\documentclass[runningheads]{llncs}
\usepackage{booktabs}
\usepackage{graphicx, subfigure}
\usepackage{dsfont}
\usepackage{amssymb}
\usepackage{mathtools}
\makeatletter
\newcommand{\printfnsymbol}[1]{%
  \textsuperscript{\@fnsymbol{#1}}%
}
\makeatother

\begin{document}
\title{How Much Can A Retailer Sell? \\Sales Forecasting on Tmall}
\author{Chaochao Chen\thanks{Equal contribution} \and Ziqi Liu\printfnsymbol{1} \and Jun Zhou, Xiaolong Li \and Yuan Qi \and Yujing Jiao \and Xingyu Zhong}
\authorrunning{Chaochao Chen et al.}
\institute{
Ant Financial Services Group \\ 
Hangzhou, China, 310099 \\
\email{\{chaochao.ccc, ziqiliu, jun.zhoujun, xl.li, yuan.qi, yujing.jyj, xingyu.zxy\}@antfin.com}\\
}

\maketitle              % typeset the header of the contribution
\begin{abstract}
Time-series forecasting is an important task in both academic and industry, which can be applied to solve many real forecasting problems like stock, water-supply, and sales predictions. 
In this paper, we study the case of retailers' sales forecasting on Tmall---the world's leading online B2C platform. 
By analyzing the data, we have two main observations, i.e., \textit{sales seasonality} after we group different groups of retails and a \textit{Tweedie distribution} after we transform the sales (target to forecast). 
Based on our observations, we design two mechanisms for sales forecasting, i.e., seasonality extraction and distribution transformation. 
First, we adopt Fourier decomposition to automatically extract the seasonalities for different categories of retailers,
which can further be used as additional features for any established regression algorithms.
Second, we propose to optimize the Tweedie loss of sales after logarithmic transformations.
%which is a Tweedie distribution instead of other losses the original ones. 
We apply these two mechanisms to classic regression models, i.e., neural network and Gradient Boosting Decision Tree, and the experimental results on Tmall dataset show that both mechanisms can significantly improve the forecasting results. 

\end{abstract}

\section{Introduction}
Time-series forecasting is an important task in both academic \cite{box2015time} and industry \cite{makridakis2000m3}, which can be applied to solve many real forecasting problems including stock, water-supply, and sales predictions. 
In this paper, we study the forecasts of retailers' future sales
at Tmall.com\footnote{\url{https://en.wikipedia.org/wiki/Tmall}},
one of the world's leading online business-to-customer (B2C) platform operated by Alibaba Group.
The problem is essentially important because accurate estimation of future sales for each retailer can help evaluate and assess the 
potential values of small businesses, and help discover potentials for further investment.

However, accurately estimating each retailer's future sales on Tmall could be
way challenging in several reasons. First of all, naturally different
goods or products are sold with strong seasonal properties. For example,
most of fans are sold in summer, while most of heaters are sold in winter,
i.e. we can observe strong seasonal properties on different groups of retailers. 
Secondly, the distribution of sales among all retailers demonstrates an over-tail power law distribution
, i.e., the retailers' sales spread a lot. 
Naively ignore such issues could make the performance much worse.

In this paper, we analyze and summarize the characteristics of the sales data at Tmall.com, and propose two mechanisms to improve forecasting performance. 
On one hand, we propose to extract seasonalities from groups of retailers. 
Specifically, we characterize the seasonal evolutions of sales by first clustering retailers into groups and then study the seasonal series as decompositions from a series of Fourier basis functions. 
The results of our approach can be utilized as features, and simply added to the feature space of any established machine learning toolkits, e.g., linear regression, neural network, and tree-based model. 
On the other hand, we place distribution transformations on the original retailers' sales. 
Specifically, we observe that the distribution over retailers' sales follows a Tweedie distribution after we transform retailers' sales by logarithmic. 
Thus, we propose to optimize Tweedie loss for regression on the logarithmic transformed sales data instead of other losses on the original ones. 
Empirically, we show that the proposed two mechanisms can significantly improve the performance of predicting 
retailers' future sales by applying them into both neural networks and tree based ensemble models.

We summarize our main contributions as follows: 
\begin{itemize}
    \item By analyzing the Tmall data, we obtain two observations, i.e., \textit{sales seasonality} after we group different categories of retailers and a \textit{Tweedie distribution} after we transform the original sales.
    \item Based on our observations, we design two general mechanisms, i.e., seasonality extraction and distribution transform, for sales forecasting. 
    Both mechanisms can be applied to most existing regression models. 
    \item We apply the proposed two mechanisms into two popular existing regression models, i.e., neural network and Gradient Boosting Decision Tree (GBDT), and the experimental results on Tmall dataset demonstrate that both mechanisms can significantly improve the forecasting results. 
\end{itemize}

\section{Data Analysis and Problem Definition}
In this section, we first describe the sales data and features on Tmall. 
We then analyze the seasonality and distribution of sales data. 
%Next, we analyze the sales data and find that it obeys Tweedie distribution after logarithmic transformations. 
Finally, we give the sales forecasting problem a formal definition. 

\subsection{Sales Data and Feature Description}\label{sec:feature}
Tmall.com is nowadays one of the largest business-to-customer (B2C) E-commerce 
platform. It has more than 180,000 retailers. 
%We show the gross merchandise volume, or GMV in figure xxx. 
Among of those retailers, there could be giant retailers like Apple.com, Prada, and together with small businesses. 
The platform is selling hundreds of thousands products in diverse categories, e.g., `furniture', `snack', and `entertainment'. 

Besides category information, the other features of retailers on Tmall can be mainly divided into three types: 
(1) The basic features that are able to reflect the marketing and selling capability of each retailer. 
For example, the amount of advertisement investment, the number of buyers, the rating/review given by the buyers, and so on. 
(2) The high-level features that are generated from historical sales and basic feature data. 
Suppose a retailer $i$ generates a series of sales data, e.g., $y_{i,t-2}, y_{i,t-1}, y_{i,t}, y_{i,t+1}$. We are currently at time $t$ and want to forecast $y_{i,t+1}$. 
Then $y_{i,t}$ can be taken as a feature which indicates the sale amount of previous period, $y_{i,t}-y_{i,t-1}$ is a feature that indicates the increasing speed of the sales, and $(y_{i,t}-y_{i,t-1}) - (y_{i,t-1} - y_{i,t-2})$ is a feature that denotes the accelerated speed of the sales. 
Similarly, we can generate other high-level features, e.g., the number of buyers, using the basic features available. 
(3) The seasonality features that are generated by using other machine learning techniques, which aim to capture the seasonal property of different retailers. 
We will present how to generate these features in Section \ref{sec:generate_season}. 

\subsection{Seasonality Analysis}\label{sec:seasonal_data}

\begin{figure}[t]
\centering 
\includegraphics[width=0.65\textwidth]{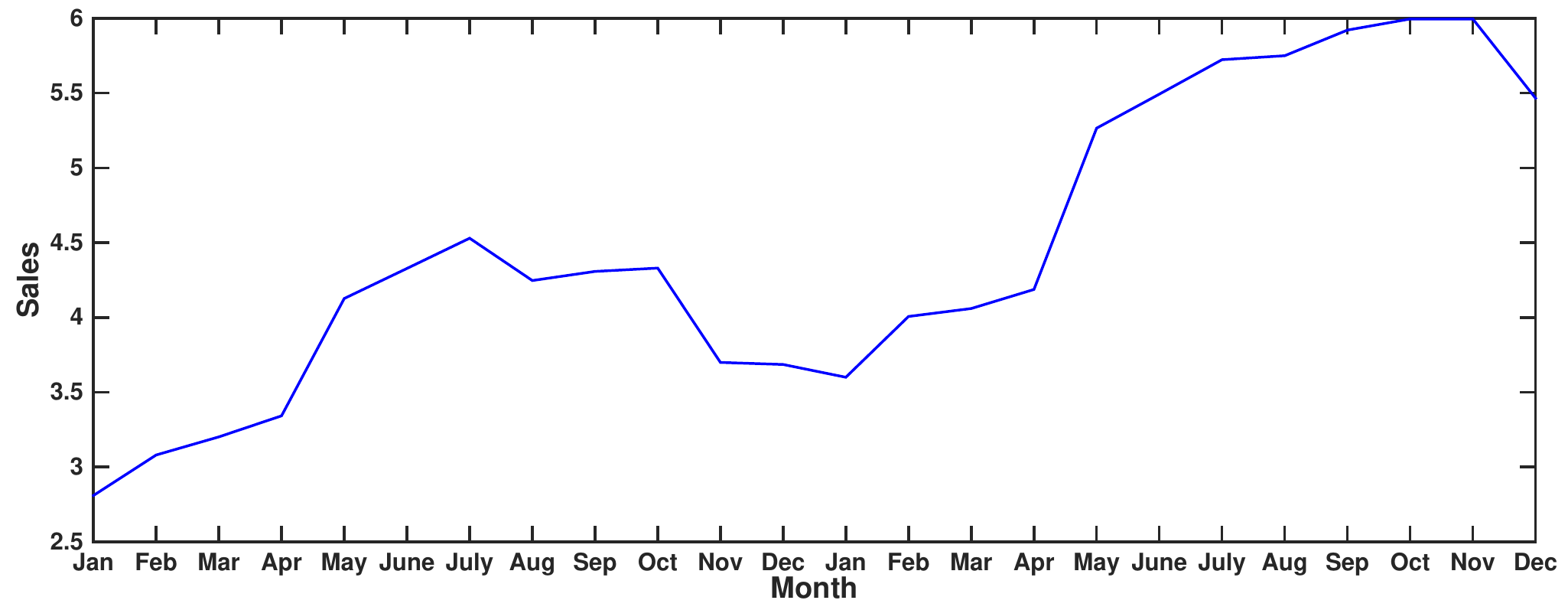} 
\vskip -0.15in
\caption{GMV on Tmall. The horizontal axis denotes the months between Jan. 2015 to Dec. 2016, and the vertical axis denotes the GMV on Tmall where we omit the scale. }
\label{fig:gmv}
\vskip -0.15in
\end{figure}

The retailers' sales tend to have different seasonality due to the seasonal items they sell. 
Take vegetables for example, tomatoes and cucumbers are usually sold more in summer, while celery cabbage is likely sold more in winter. 
Although the GMV demonstrate seasonal properties, as is shown in Figure \ref{fig:gmv}, the analysis of the seasonality related to the gross merchandise volume is relative meaningless for the prediction on each retailer. 
In contrast, the seasonality analysis on each single retailer makes the analysis cannot generalize well in the future. 
Instead, we further investigate the seasonalities in different groups of retailers. 

\begin{figure*}[t]
  \centering 
  \subfigure[category=`Women's Wearing'] { \includegraphics[width=0.45\textwidth]{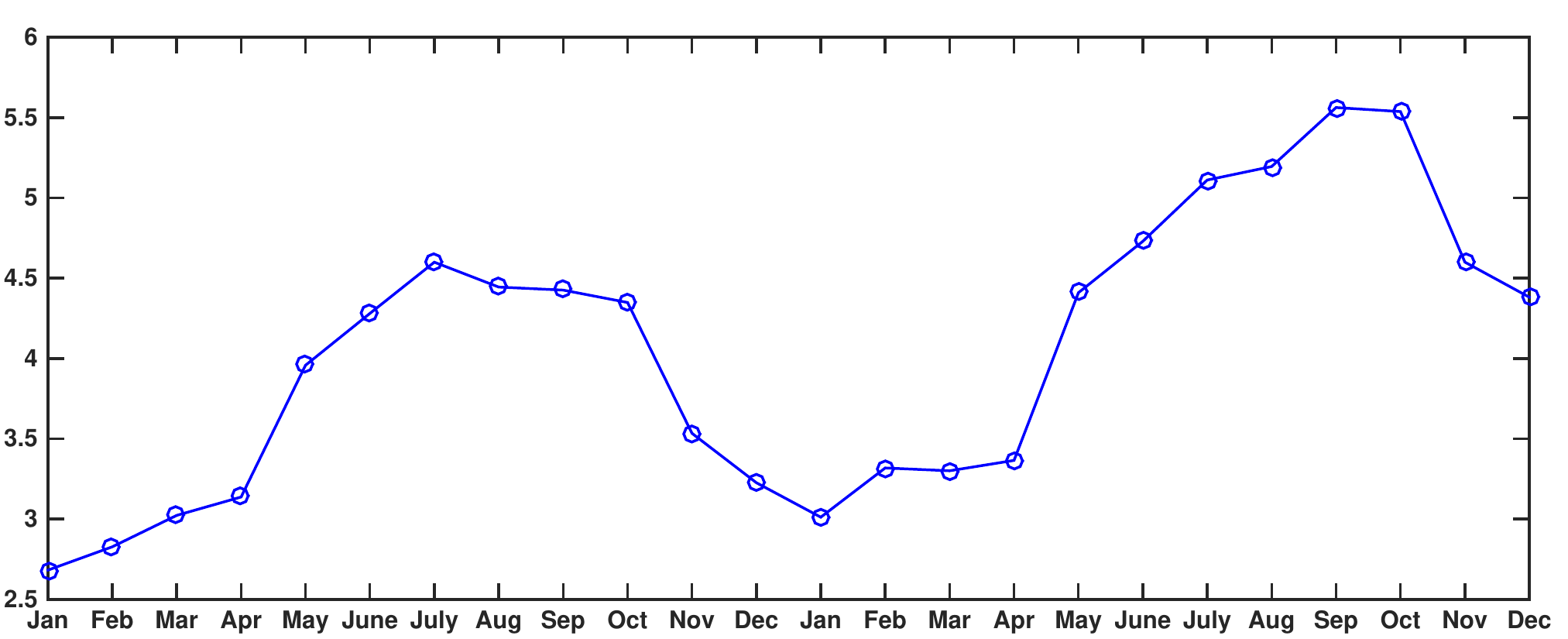} }
   \subfigure[category=`Men's Wearing'] {  \includegraphics[width=0.45\textwidth]{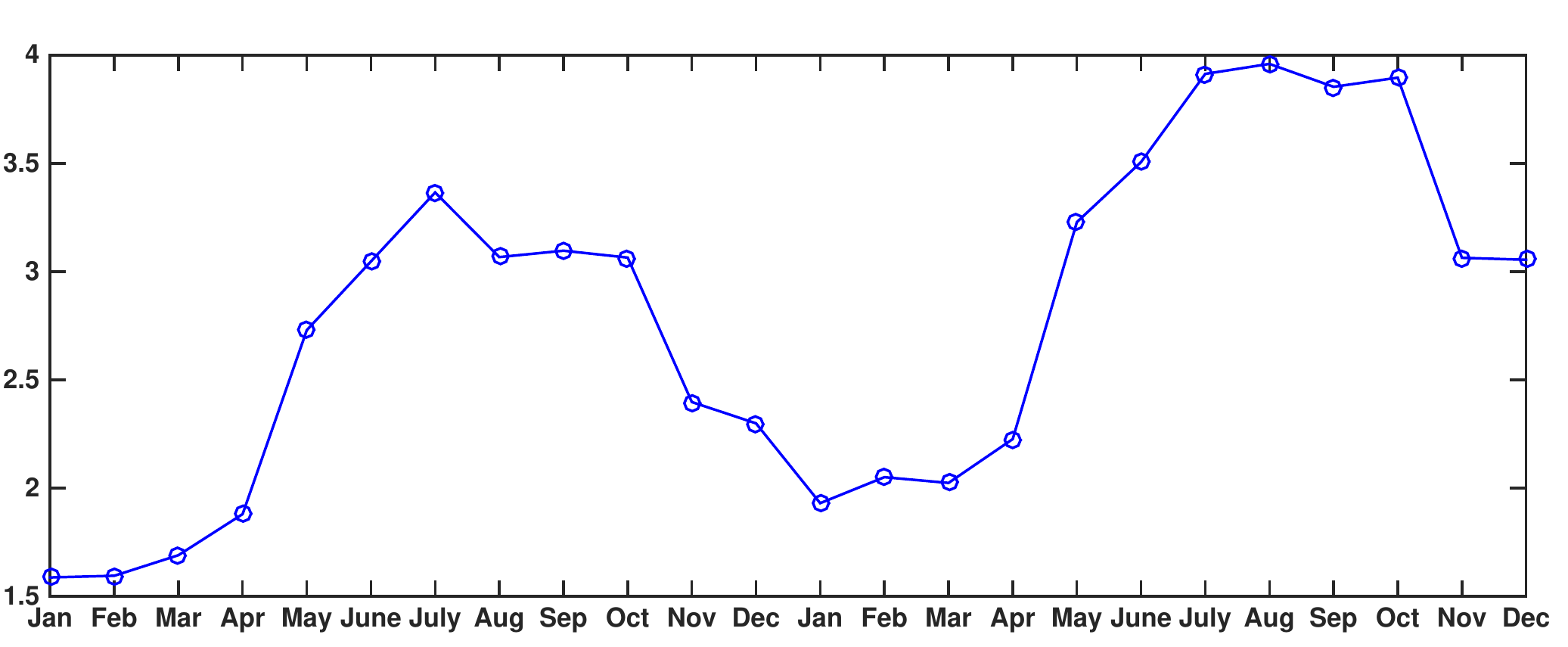} }
  \subfigure[category=`Snack'] {   \includegraphics[width=0.45\textwidth]{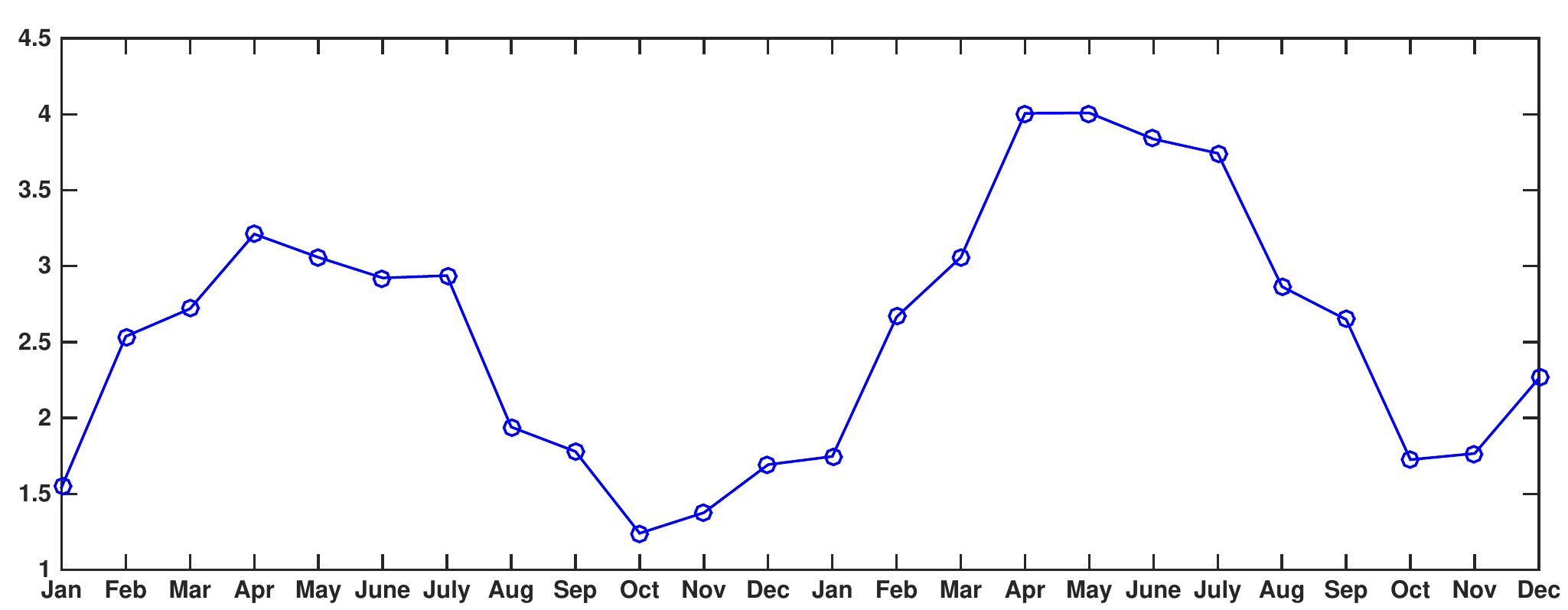} }
  \subfigure[category=`Meat'] {   \includegraphics[width=0.45\textwidth]{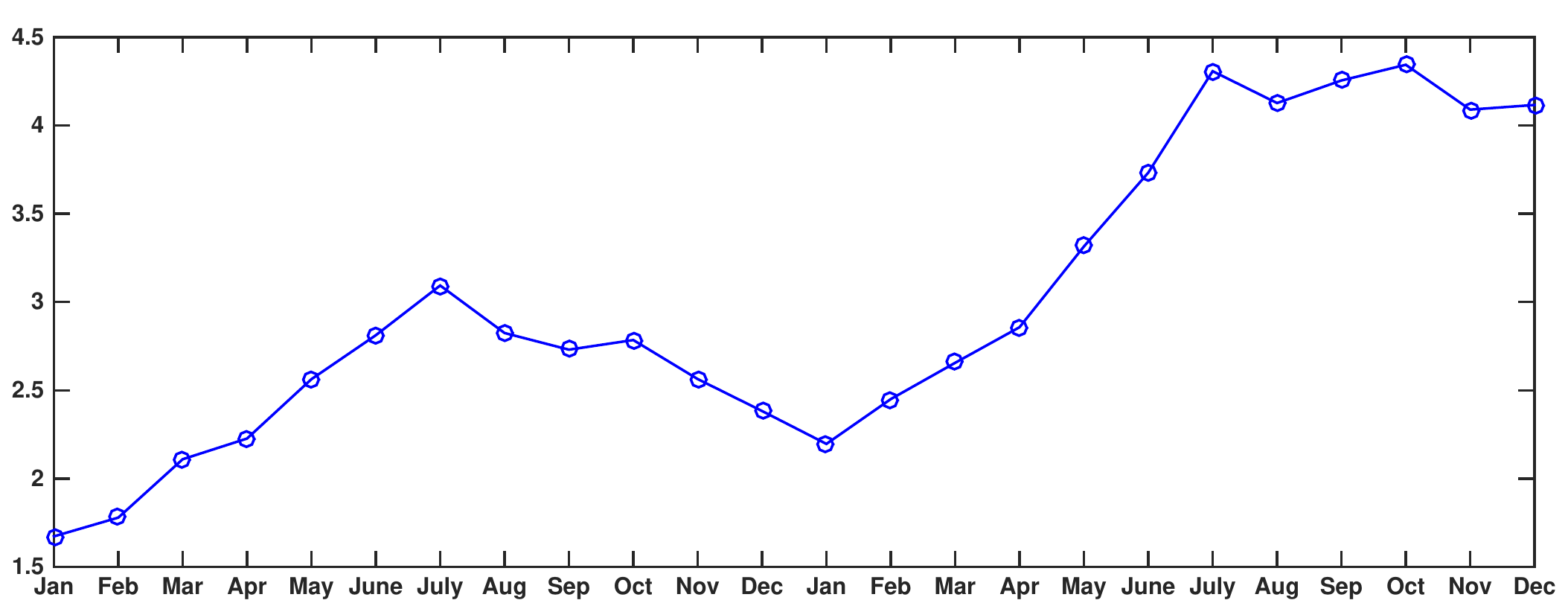} }
  \vskip -0.15in
    \caption{Sales seasonality of different categories on Tmall. The horizontal axis denotes the months between Jan. 2015 to Dec. 2016, and the vertical axis denotes the total sale among of retailers in each category where we omit the scale. }
    \label{fig:season}
    \vskip -0.15in
\end{figure*}

By analyzing the sales data on Tmall, we observe different seasonal patterns on different categories of retailers. 
Figure \ref{fig:season} shows the sales of four different categories of retailers, i.e., `Women's Wearing', `Men's Wearing', `Snack', and `Meat', where we use two year's sales data from January 2015 to December 2016. 
We can observe that, `Women's Wearing' and `Men's Wearing' show quite similar seasonal patterns, i.e., they both reach peak in summer (July or August) and decline to nadir in winter (January). 
On the contrary, `Snack' and `Meat' show different seasonal patterns. 
In summary, the seasonalities under different categories could differ quite a lot. 
Thus if we can somehow partition the retailers into appropriate groups, the shared seasonality among retailers in one group could be statistically useful for characterizing each retailer in the group. 
Given a group of retailers, how can we characterize
the seasonality for the group remains to be solved.
We will discuss our approaches in Section 3.1. 

\subsection{Sale Amount Analysis}\label{sec:sales_data}
\begin{figure*}[t]
  \centering 
  \includegraphics[width=0.45\textwidth]{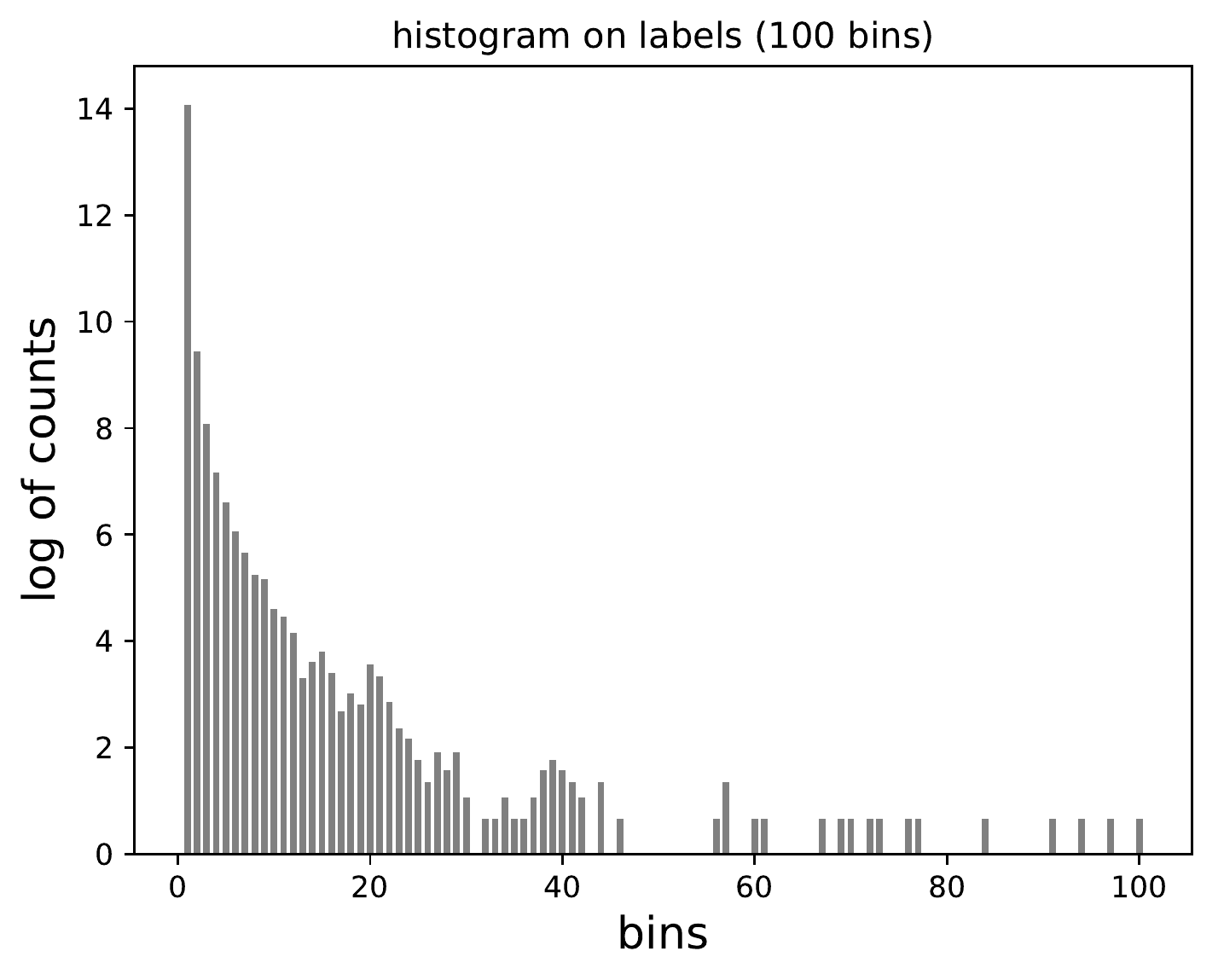}
    \includegraphics[width=0.45\textwidth]{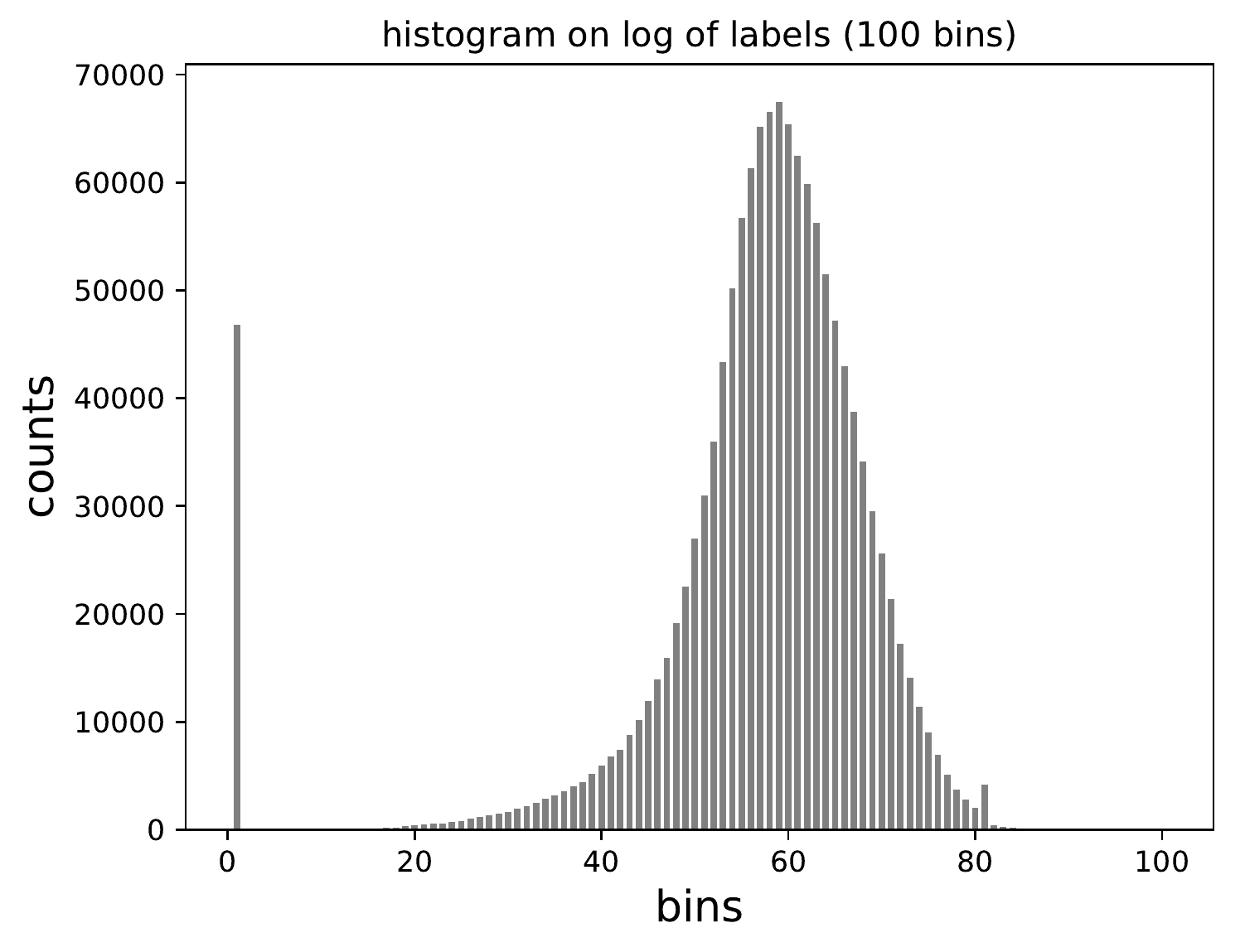}
    \vskip -0.15in
    \caption{Sales on Tmall obey Tweedie distribution after logarithmic transformation. We order the sales among retailers in an increasing order, partition the sales into 100 bins with equal frequency binning (x-axis), and show the retailer counts in each bin (y-axis). }
    \label{fig:label}
    \vskip -0.15in
\end{figure*}

The sales of each retailer over time-series could be much challenging. 
To illustrate this, we show the histograms over sales in Figure~\ref{fig:label} (left). It 
shows that the sales could be much diverse across over all the retailers. In practice,
this is very hard to formalize as a trivial regression problem because the errors
on those sales from giant retailers could dominate the loss, e.g. least squared loss. 

Instead, after we do a logarithmic transformation on the sales of each retailer, we
found that the histogram appears to be a clear Tweedie distribution, i.e. Figure~\ref{fig:label} (right), which will be further described in details in Section \ref{sec:tweedie_trans}.
As a result, such transformation on the dependent variables makes our forecasting much easier.
Note that, there are always some retailers' sale around zero. 
This is because some shops on Tmall will close or forced to be closed by Tmall due to some reason from time to time, and correspondingly, some shop will be newly opened or reopen. 
Consequently, some retailers' sales are around zero. 

\subsection{Problem Definition}
Assuming any retailer in Tmall.com as $i$, at month $t$, We formalize the sales forecasts problem as a regression problem, i.e. given the features of each retailer $x_{i,< t} \in \mathbb{R}^{d}$, where $d$ is the feature dimensionality and $< t$ denotes the months before $t$, and the known sales of each retailer $y_{i,< t+1}$, we want to learn a function: $f: x_{i,t} \mapsto y_{i,t+1}$ , where $x_{i,t}$ denotes the features of  retailers $i$ at month $t$ and $y_{i,t+1}$ denotes its sales at month $t+1$. % to the $t+6$-th month.  
That is, give the features of any retailer at month $t$, we want to predict their corresponding sales at month $t+1$. 
%The reason why we aim to predict the next 6 months sales is based on a consideration of credit strategy at Alibaba. From now on, by default, the sales will denote a retailer's average sales of the next 6 months.
\section{Model Design and implementation}
In this section, we will present our designed two mechanisms, i.e., seasonality extraction and distribution transform, for sales forecasting.

\subsection{Seasonality Over Groups of Retailers}\label{sec:generate_season}
As we reported in Section \ref{sec:seasonal_data}, the seasonalities under different categories could differ quite a lot, therefore, the remaining problem is that how should we partition the retailers into appropriate groups. 
Instead of manually partition retailers, we adopt clustering methods for time-series data \cite{liao2005clustering} to do so.
Specifically, we group the retailers by using the basic and high-level features described in Section \ref{sec:feature},  so that retailers that have similar features are grouped together.

After we partition retailers into groups, we adopt discrete Fourier transform to automatically extract the seasonality for retailers in different groups.  
Formally, assuming a group of sellers with expected amount of sales annotated as $\tilde{y}(t)$ at time $t$,
thus results into a series of expected sales as observations, i.e. $\{\tilde{y}(0), ..., \tilde{y}(t),..., \tilde{y}(T)\}$.
Each periodic function $\tilde{y}(\cdot)$ can be expanded by the Fourier series, which is a linear combination
of infinite sines and cosines,
\begin{align}\label{eq:fourier}
\tilde{y}(t) = a_0 + \sum_{n=1}^{\infty} a_n \mathrm{cos}(n t) \sum_{n=1}^{\infty} b_n \mathrm{sin}(n t),
\end{align}
where $\big\{a_i, b_i | i \in \{0,..., n, ..., \infty\}\big\}$ are parameters to be optimized.
As a result, the function $\tilde{y}(\cdot)$ can be represented by a Fourier basis.

\begin{figure*}[t]
\centering 
\includegraphics[width=0.4\textwidth]{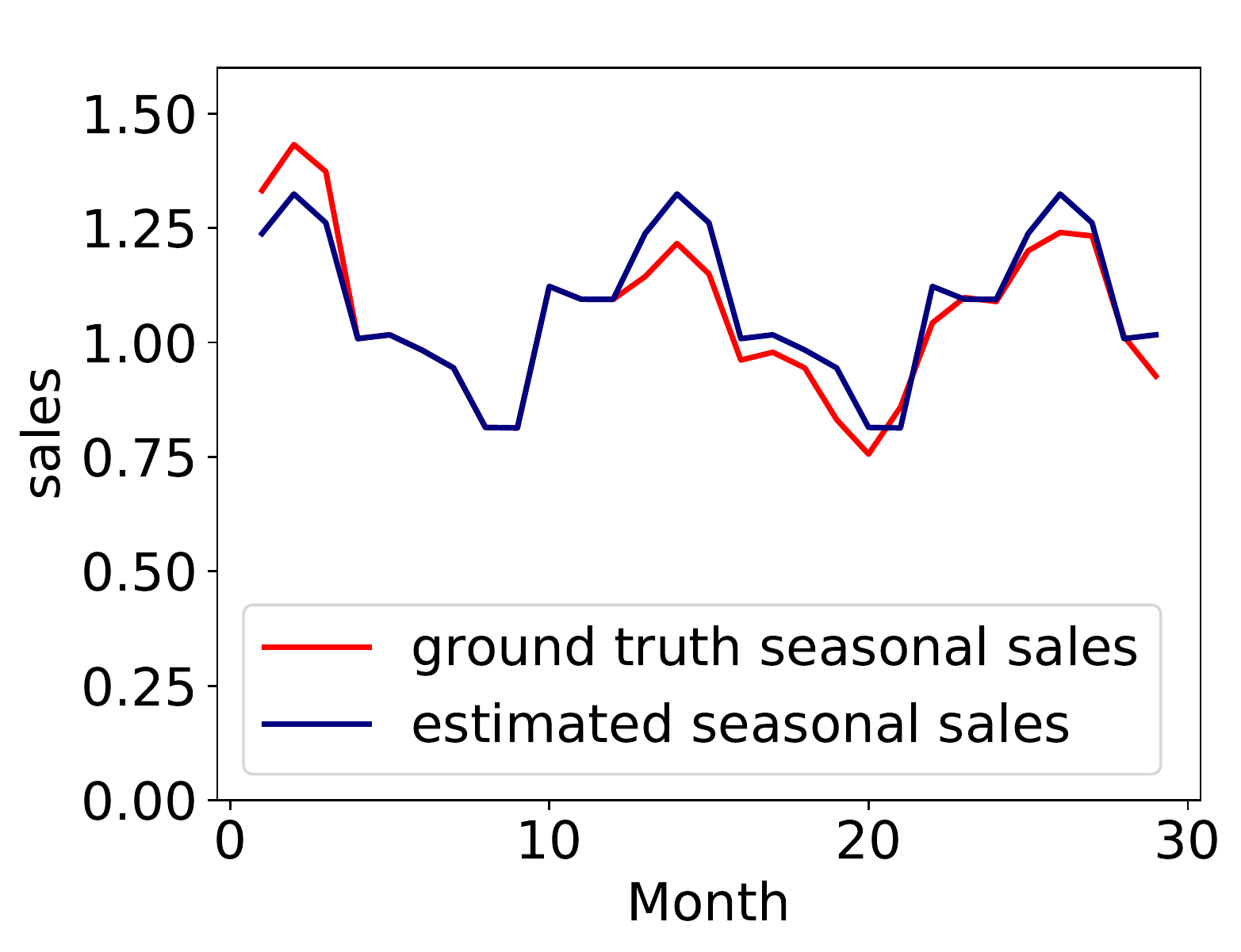} 
\includegraphics[width=0.4\textwidth]{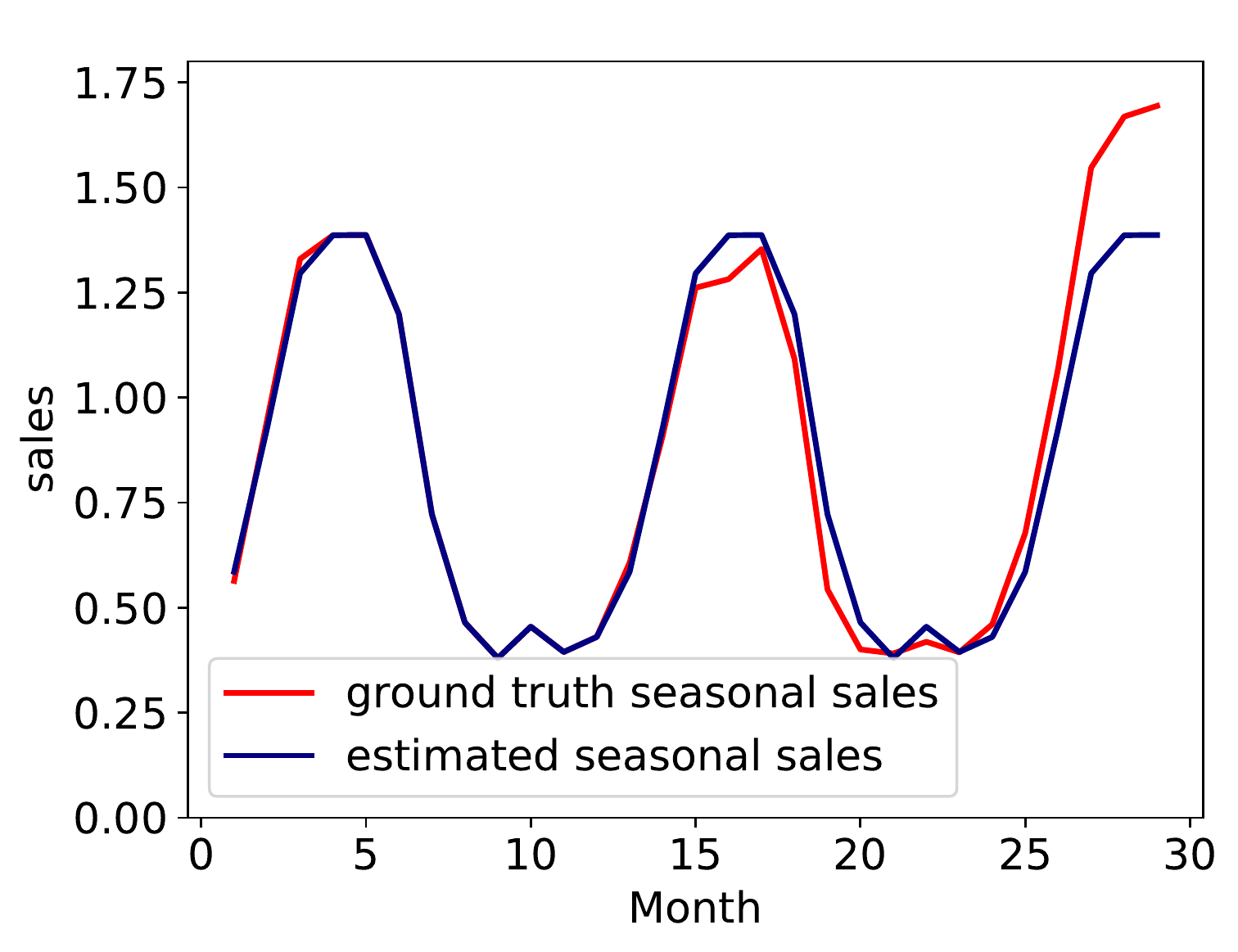} 
\vskip -0.15in
\caption{Seasonality extraction results for two groups of retailers. The left one is mainly the group of retailers who sell purses, and the right one is mainly the group of retailers who sell accessaries. In both figures, we use the first 15 months data to learn the parameters in Eq.(\ref{eq:fourier}), and further use them to predict the seasonality of all the months' data (note that we also omit the scale of the sale amount). }
\vskip -0.15in
\label{fig:fourier}
\end{figure*}

We now show the results of extracted seasonalities on different groups of retailers. 
We randomly select two groups of retailers and show their seasonalities and estimates for sales in Figure~\ref{fig:fourier}, where we find the two groups of retailers mainly sell purses and accessaries, respectively. 
In Figure~\ref{fig:fourier}, we use the first 15 months' data to learn the parameters in Eq.(\ref{eq:fourier}), and further use them to predict the seasonality of all the months' data. 
It is obvious that our extracted seasonality is very close to the real one in both groups. 
Similarly, we can extract seasonality for other features, e.g., the number of buyers and the among of advertisement investment, by using the same method. 

In practice, we use two types of features extracted from such seasonal patterns: 
(1) the seasonal values of the target we want the extract, e.g., sales and the number of buyers, in a window of 12 months centered around the month $t$. 
(2) the variation, i.e. the difference among those seasonal values. 
Hopefully, such seasonality or trend measures for each group of sellers can be fed into any
classifiers, so as to characterize the seasonal patterns for each seller. 
We will empirically study the effectiveness of these seasonality features in experiments. 

\subsection{Tweedie Loss For Regression}\label{sec:tweedie_trans}

As we described in Section \ref{sec:sales_data}, based on our observation, the sales on Tmall will clearly obey Tweedie distribution after a logarithmic transformation. 
From Figure~\ref{fig:label} (right), we see that the sales after logarithmic transformation is a combination of Poisson distribution and Gamma distribution, which is a special case of Tweedie distribution, i.e., a compound Poisson-Gamma distribution. %when $1 < p < 2$ ($p$ will be explained later). 
That is, we assume that (1) the status of retailers, i.e., closed or not, are independent identically distributed and they obey Poisson distribution; 
(2) the sales of retailers are also independent identically distributed and they obey Gamma distribution. 
The Tweedie distribution was first proposed in \cite{tweedie1984index}, and then officially named by Bent Jørgensen in \cite{jorgensen1987exponential}, which belongs to the class of exponential dispersion. 

Tweedie distribution has been popularly used in insurance scenarios \cite{yang2017insurance}. 
We now formally describe Tweedie distribution in sale forecasting scenario. 
Suppose %$y \sim Poisson(\lambda)$ and $y \sim Gamma(\alpha, \theta)$ 
Let $N$ be a Poisson random variable denoted by $Pois(\lambda)$, and let $Z_i, i=0,1,2,...,N$ be independent identically distributed gamma random variables denoted by $Gamma(\alpha,\gamma)$ with mean $\alpha\gamma$ and variance $\alpha\gamma^2$. 
We also assume that $N$ is s independent of $Z_i$. 
Define a random variable $Z$ by
\begin{equation}\label{eq:td}
\begin{split}
Z = 
\begin{cases}
0,& \text{if $N=0$},\\
\text{$Z_1+Z_2+...+Z_N$} ,& \text{if $N>0$}.
\end{cases}
\end{split}
\end{equation} 

We can see from Eq.(\ref{eq:td}) that $Z$ is the Poisson sum of independent Gamma random variables, which is also called compound Poisson-Gamma distribution. 
In sales forecasting scenarios, $Z$ can be viewed as the total number of retailers, $N$ as the opened retailers, and $Z_i$ as the sale amount of retailers $i$. 
Note that the distribution of $Z$ has a probability mass at zero, i.e., $Pr(Z = 0) = exp(-\lambda)$.
The existing research has proven that, if we reparametrize $(\lambda, \alpha, \gamma)$ by 
\begin{equation}\nonumber
  \lambda = \frac{1}{\phi} \frac{\mu^{2-p}}{2-p}, ~~~~~
  \alpha = \frac{2-p}{1-p}, ~~~~~
  \gamma = \phi (\rho-1)\mu^{\rho-1},
\end{equation} 
Eq.(\ref{eq:td}) then becomes the form of a Tweedie model $Tw(\mu,\phi,\rho)$ with $1<\rho<2$ and $\mu>0$.
Here, the boundary cases $\rho \rightarrow 1$ and $\rho \rightarrow 2$  correspond to the Poisson and the gamma distributions, respectively. 
The compound Poisson-Gamma distribution with $1 < \rho < 2$ can be seen as a bridge between the Poisson and the Gamma distributions.

%We also assume that the number of retailers are realizations of random variables $R$, i.e., the number of retailers and the among of their sales are both modeled stochastically. This can be done by assuming that $R$ is Poisson distributed. This leads to the following compound Poisson-gamma distribution which is parametrized by three parameter $\lambda$, $\tau$, and $\gamma$, 

%\begin{equation}
% f(y|p, \mu, \phi) = c(y|p, \phi) \cdot exp \left(\frac{1}{\phi} \left( \frac{y \cdot \mu^{1-p}}{1-p} - \frac{\mu^{2-p}}{2-p} \right) \right),
%\end{equation} 
%where 
%\begin{equation}\nonumber
%\begin{split}
% & c(y|p, \phi) = \sum_{r}{\left\{\frac{y^r}{\phi^{r+1}(p-1)^\gamma(2-p)} \frac{1}{r!  \tau(r\gamma)y} \right\}},\\
% & p = \frac{\alpha+2}{\alpha+1} \in (1,2), \\
% & \mu = \lambda \cdot \tau, \\
% & \phi = \frac{\lambda^{1-p} \cdot \tau ^ {2-p}}{2-p}.
%\end{split}
%\end{equation} 

%After we set the parameter for Tweedie distribution, when we want to perform a regression task. 
The log-likelihood of this Tweedie model for the sale $y$ of a retailer is
\begin{equation}
  L(y|\mu, \phi, \rho) = \frac{1}{\phi}\left(y \frac{1}{\phi} \frac{\mu^{1-p}}{1-p} - \frac{1}{\phi} \frac{\mu^{2-p}}{2-p} \right)
        + log(a(y, \phi, \rho)),
\end{equation} 
%\begin{equation}
%\begin{split}
% L(p, \mu, \phi)  & = \sum_{i}[ r_i log \left\{ \frac{y_i^r}{\phi^{r+1}(p-1)^\gamma(2-p)}  \right\} - log ( {r_i!  \tau(r_i \gamma)y_i} ) \\
% & + \frac{1}{\phi}\left\{ y_i \frac{\mu_i^{1-p}}{1-p} - \frac{\mu_i^{2-p}}{2-p} \right\} ].
%\end{split}
%\end{equation} 
where the normalizing function $a(\cdot)$ can be written as
\begin{equation}\nonumber
\begin{split}
a(y, \phi, \rho) = 
\begin{cases}
\text{$\frac{1}{y}\sum_{t} W_t(y, \phi,\rho)$},& \text{for $y>0$},\\
1 , & \text{for $y=0$},
\end{cases}
\end{split}
\end{equation} 
and $\sum_{t} W_t$ is an example of Wright's generalized Bessel function \cite{tweedie1984index}. 

After that, given the parameter $\rho$ for Tweedie model, the other parameters can be efficiently solved by using maximum log-likelihood approach \cite{yang2017insurance}. 
The Tweedie model can be naturally combined with most existing regression models, e.g., NN and GBDT. 
That is, we can train a Tweedie  loss NN model or GBDT model instead of the models with other losses, e.g., square loss \cite{yang2017insurance}. 
Obviously, the results of Tweedie loss regression are much better than those of other loss regression, e.g., square loss, as will be shown in experiments. 
This is because Tweedie loss fits the real sales distribution after logarithmic transformation of sales, as is shown in Figure \ref{fig:label} (right).  

\section{Empirical Study}
In this section, we first describe the dataset and the experimental settings. %, including the evaluation metric and comparison methods. 
Then we report the experimental result by comparing with various state-of-the-art sales forecasting techniques. 
We finally analyze the effect of Tweedie distribution parameter ($\rho$) on model performance. 

\subsection{Dataset}

\noindent \textbf{Features}. 
As we described in Section 2, the features of retailers mainly contain three
types, i.e., the basic features, high-level features that are generated from
historical sales and basic feature data,  and the seasonality features that
are generated by using other machine learning techniques. 
%Sales forecasting on Tmall can help evaluate and assess the potential
%values of small businesses, and further help discover potentials for further investment. 
%Thus, the stability of the forecasting model is important. 
%To make sure of it, before deployment,
%we use the Population Stability Index (PSI)---an important metric to identify
%a shift in population for retail credit scorecards---to select stable features. 
This includes 189 features in total, where there are 79 basic features,
102 high-level features, and 8 seasonality features as we discussed in
section~\ref{sec:generate_season}. 

\noindent \textbf{Samples}. 
We choose the samples (retailers) during Jan. 2015 and Dec. 2016 on Tmall. 
Note that we only focus on forecasting the relative small retailers whose monthly sale amount is under a certain range (300,000). 
Because, in practice, the sales of big retailers are very stable, and it is meaningless to forecast their sales. 
After that, we have 783,340 samples. 
%In our data, we only choose the retailers whose monthly sales are under 300,000. 
We use the samples in 2015 as training data, the samples from Jan. 2016 to June 2016 as validation, and the samples from July 2016 to Dec. 2016 as test data. 
%We use the validate data to tune parameters and use test data to evaluate model performance. 
%The statistical of the dataset is summarized in Table \ref{dataset}. 

%\begin{table}
%\centering
%\caption{Dataset description}
%%\vskip -0.15in
%\label{dataset}
%\begin{tabular}{|c|c|c|c|}
%  \hline
%  Dataset & \#retailer & \#level & \#feature  \\
%  \hline
%  \hline
%  \emph{Netflix} & xxx & xxx & xxx \\
%  \hline
%\end{tabular}
%%\vskip -0.15in
%\end{table}

\subsection{Experimental Settings}

\noindent \textbf{Evaluation metric}. 
Most existing research use error-based metric, e.g., Mean Average Error (MAE) and Root Mean Square Error (RSME), to evaluate the performance in time-series forecasting \cite{carbonneau2008application,ahmed2010empirical}. 
However, these metrics are way sensitive to those retailers whose sales are large. 
As we can see in Figure 1, the sales on Tmall spread a lot. 
In practice, the forecasting precision of the retailers with small sales counts the same as the ones with big sales. 
Therefore, we propose to use Relative Precision (RP) for sales forecasting on Tmall, which is defined as
\begin{equation}\label{eq:metric}
	RP@p=\frac{\sum_{i=1}^{N} \mathds{1} \left( \frac {|y_i - \hat{y}_i|}{y_i} <  p  \right)}{N},
\end{equation} 
where $N$ is the total number of retailers, $y_i$ as the real sale and $\hat{y}_i$ as the forecasted sale, $p \in [0, 1]$, and $\mathds{1}(\cdot)$ is the indicator function that equals to 1 if the expression in it is true and 0 otherwise. 

As we can see from Eq. (\ref{eq:metric}), RP is actually the percentage of the retailers whose forecasting error is in a certain range $p$.  
Intuitively, the smaller $p$ is, the smaller RP will be. 
Because one has higher demanding for the forecasting performance when $p$ is smaller. 

%Following the existing time-series forecasting research \cite{carbonneau2008application,ahmed2010empirical}, we use Mean Average Error (MAE) and Symmetric Mean Absolute Percentage Error (SMAPE) as the evaluation metrics. 
%MAE is commonly used as the metric to evaluate the performance of regression tasks, and SMAPE can reduce the influence of these low sale retailers and thus is popularly used in time-series forecasting problems. They are defined as below:
%
%\begin{equation}\nonumber
%\begin{split}
%	MAE=\frac{\sum_{i=1}^{N}|y_i - \hat{y}_i|}{N},\\
%	SMAPE=\frac{2}{N}\sum_{i=1}^{N} \frac{|y_i - \hat{y}_i|}{y_i + \hat{y}_i},
%  \end{split}
%\end{equation} 
%where $N$ denotes the total number of samples (retailers) in the test set, and $y_i$ and $\hat{y}_i$ denote the real sales and the predicted sales for each retailer respectively. 

\noindent \textbf{Comparison methods}. 
Our proposed mechanisms, i.e., seasonality extraction and distribution transform, has the ability to generalize to most existing regression algorithms. 
To prove this, we apply the mechanisms into two popular regression models, i.e., Neural Network (NN) and Gradient Boosting Decision Tree (GBDT). 
We summarize all the methods, including ours, as follow:

\begin{itemize} 
    \item \textbf{NN} has been used to do time-series forecasting and proven effective where we use square loss \cite{qi2008trend,ahmed2010empirical}. 
    %Specifically, we use a three-layer network, with Rectified Linear Unit (ReLU) as active function, learning rate of 0.1 and a momentum of 0.7. 
    \item \textbf{NN-S} uses extra our proposed seasonal feature in Section \ref{sec:generate_season} for NN, and its comparison with NN will prove the effectiveness of seasonality extraction. 
    \item \textbf{NN-T} uses our proposed Tweedie-loss in Section \ref{sec:tweedie_trans} for NN, and its comparison with NN will prove the effectiveness of our proposed Tweedie-loss regression after sale distribution transform. 
    \item \textbf{NN-ST} extra uses our proposed seasonal feature in Section \ref{sec:generate_season} for NN-T, which is the application of our proposed two mechanisms in NN.
    \item \textbf{GBDT} is developed for additive expansions based on any fitting criterion, which belongs to a general gradient-descent `boosting' paradigm and suits for regression tasks with many types of loss functions, e.g., least-square loss, Huber loss, and Tweedie loss \cite{friedman2001greedy}. Specifically, we use the GBDT algorithms implemented on Kunpeng \cite{zhou2017kunpeng}---a distributed learning system that is popularly used in Alibaba and Ant Financial, where we also use square loss. 
    \item \textbf{GBDT-S} uses extra seasonal feature for GBDT, similar as NN-S. 
    \item \textbf{GBDT-T} uses Tweedie-loss for GBDT, similar as NN-T. 
    \item \textbf{GBDT-ST} uses extra seasonal feature for GBDT-T, similar as NN-ST.
\end{itemize}

\noindent \textbf{Parameter setting}. For NN, we use a three-layer network, with
Rectified Linear Unit (ReLU) as active functions, and optimized with Adam \cite{kingma2014adam} (learning rate as 0.1). 
For GBDT, we set tree number as 120, learning rate as 0.3, and regularizer of $\ell_2$ norm as 0.5. 
We will study the effect of parameter $\rho$ of Tweedie regression in Section 4.4. 

\subsection{Comparison Results}

\begin{table}[t]\label{compare}
\centering
\caption{Comparison result on test data.}
\vskip -0.1in
\begin{tabular}{|c||cccc||cccc|}
  \hline
  Model & NN  & NN-S  & NN-T & \textbf{NN-ST} & GBDT  & GBDT-S  & GBDT-T & GBDT-ST \\
  \hline
  \hline
  RP@0.1 & 0.1693 & 0.1723 & 0.3236 & \textbf{0.3338} & 0.1719 & 0.1859 & 0.3159 & \textbf{0.3263} \\
  \hline
  RP@0.2 & 0.1933 & 0.1987 & 0.3484 & \textbf{0.3534} & 0.2095 & 0.2242 & 0.3394 & \textbf{0.3520} \\
  \hline
  RP@0.3 & 0.2603 & 0.2657 & 0.3950 & \textbf{0.3956} & 0.2681 & 0.2816 & 0.3821 & \textbf{0.3966} \\
  \hline
\end{tabular}
\vskip -0.15in
\end{table}

We summarize the comparison results in Table 1, and have the following comments.
%\begin{itemize} 
%    \item The forecasting performance of NN and GBDT are close, and the performance of GBDT is slightly higher than NN. 
%    This is because GBDT can naturally consider the complicate relationship, e.g., cross feature, between features. 
%    \item Our proposed seasonality extraction mechanism can clearly improve the forecasting performance of both NN and GBDT. 
%    For example, GBDT-S improves the forecasting performance of GBDT by 8.14\% in terms of RP@0.1, and GBDT-ST further improves the forecasting performance of GBDT-T by 3.29\%.
%    \item Our proposed distribution transform mechanism can significantly improve the forecasting performance of both NN and GBDT. 
%    For example, NN-T improves the forecasting performance of NN by 91.14\% in terms of RP@0.1, and NN-ST improves the forecasting performance of NN-S by 93.73\% in terms of RP@0.1
%    \item In summary, our proposed two mechanisms consistently improve the forecasting performances of both NN and GBDT models. 
%    Specifically, NN-ST improves the forecasting performance of NN by 97.14\%, 82.82\%, 51.98\% in terms of RP@0.1, RP@0.2, and RP@0.3 respectively. 
%    And, GBDT-ST improves the forecasting performance of GBDT by 89.82\%, 68.10\%, 47.93\% in terms of RP@0.1, RP@0.2, and RP@0.3 respectively. 
%    The results not only demonstrate the effectiveness of our proposed mechanisms, but also indicate the generalizability of them. 
%\end{itemize}

    \textbf{(1)} The forecasting performance of NN and GBDT are close, and the performance of GBDT is slightly higher than NN. 
    This is because GBDT can naturally consider the complicate relationship, e.g., cross feature, between features. 
    \textbf{(2)} Our proposed seasonality extraction mechanism can clearly improve the forecasting performance of both NN and GBDT. 
    For example, GBDT-S improves the forecasting performance of GBDT by 8.14\% in terms of RP@0.1, and GBDT-ST further improves the forecasting performance of GBDT-T by 3.29\%.
    \textbf{(3)} Our proposed distribution transform mechanism can significantly improve the forecasting performance of both NN and GBDT. 
    For example, NN-T improves the forecasting performance of NN by 91.14\% in terms of RP@0.1, and NN-ST improves the forecasting performance of NN-S by 93.73\% in terms of RP@0.1
    \textbf{(4)} In summary, our proposed two mechanisms consistently improve the forecasting performances of both NN and GBDT models. 
    Specifically, NN-ST improves the forecasting performance of NN by 97.14\%, 82.82\%, 51.98\% in terms of RP@0.1, RP@0.2, and RP@0.3 respectively. 
    And, GBDT-ST improves the forecasting performance of GBDT by 89.82\%, 68.10\%, 47.93\% in terms of RP@0.1, RP@0.2, and RP@0.3 respectively. 
    The results not only demonstrate the effectiveness of our proposed mechanisms, but also indicate the generalizability of them. 
\begin{figure}[t]
\centering 
\includegraphics[width=0.35\textwidth]{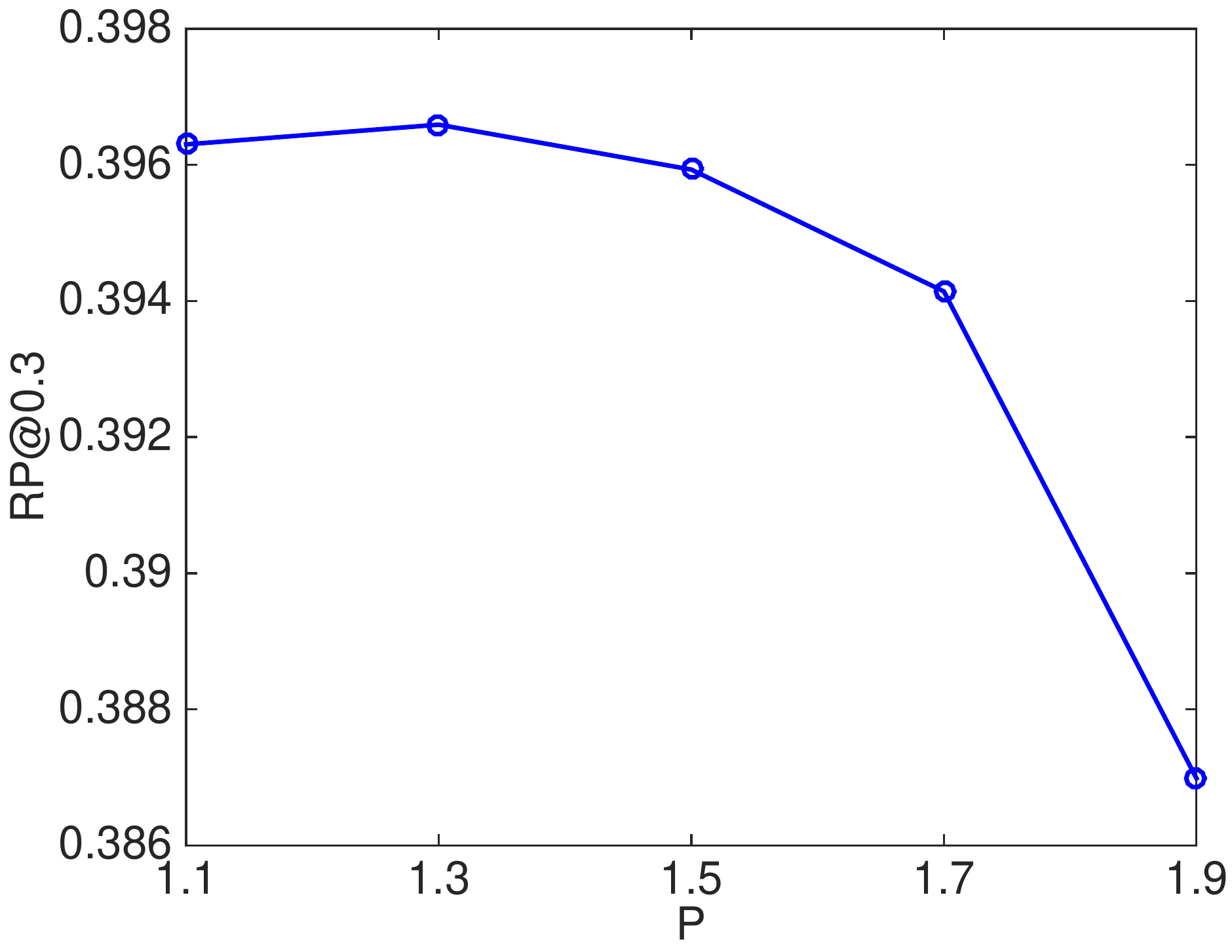} 
\vskip -0.15in
\caption{Effect of Tweedie distribution parameter ($\rho$) on GBDT-ST on validate data. }
\label{fig:effect}
\vskip -0.15in
\end{figure}

\subsection{Effect of Tweedie distribution parameter ($\rho$) }
As described in Section 3.2, the Tweedie distribution parameter ($\rho$)  bridges  the Poisson and the Gamma distributions, and the boundary cases $\rho \rightarrow 1$ and $\rho \rightarrow 2$  correspond to the Poisson and the Gamma distributions, respectively. 
The effect of Tweedie distribution parameter ($\rho$) on GBDT-ST is shown in Figure \ref{fig:effect}, where we use the validate data. 
From it, we find that GBDT-ST achieves the best performance when $\rho=1.3$. 
This indicates that the real sales data on Tmall fit the Tweedie distribution when $\rho=1.3$. 
%In practice, the best value of $p$ can be easily find using 

%\subsection{Deployment}
%Our proposed time-series forecasting methods have been successfully deployed for sales forecasting on Tmall. 
%We finally choose GBDT-ST for deployment based on two reasons: 
%First, tree based model is more stable than neural network. 
%Second and the most important, tree based model has better explanation ability than neural network. 
%In practice, our sales forecasting model can be used to assess the potential of retailers, and further used to make investment on them. 
%We have to show the certain reasons why retailers can not get investment if they ask for it. 
%Therefore, a good explanation ability is essential. 
%Specifically, our model is first trained and evaluated, and then deployed online. 
%Finally, the forecasting model will be triggered at the end of each month by using the known features, and its output are the forecasted sale of each retailer in the next month. 

\section{Related Works}
In this section, we will review literatures on time-series forecasting, which are mainly in two types, i.e., linear model and non-liner model. 

\subsection{Linear Model}
The most popular linear models for time-series forecasting are linear regression and Autoregressive Integrated Moving Average model (ARIMA) \cite{hannan2009multiple}. 
Due to their efficiency and stability, they have been applied to many forecasting problems, e.g., wind speed \cite{kavasseri2009day}, traffic \cite{sun2003use}, air pollution index \cite{lee2012seasonal}, electricity price\cite{bianco2009electricity}, and Inflation \cite{pufnik2006short}. 
However, since it is difficult for them to consider complicate relations between features, e.g., cross feature, their performance are limited. 

\subsection{Non-Linear Model}
Non-linear models are also adopted for time-series forecasting. 
The most popular ones are Support Vector Machine (SVM), neural network, and tree-based ensemble models. 
For example, SVM are applied to financial forecasting \cite{kim2003financial} and wind speed forecasting \cite{liu2014short}. 
Neural network are also used in financial marketing forecasting \cite{azoff1994neural} and electric load forecasting  \cite{park1991electric}. 
Recently, Gradient Boosting Decision Tree (GBDT) are also adopted to forecast traffic flow \cite{yinga2017traffic}. 

Moreover, model ensemble is also popular for time-series forecasting. 
For example, ARIMA and SVM were combined to forecast stock price \cite{pai2005hybrid}. 
Hybrid ARIMA and NN models were also used for time-series forecasting \cite{zhang2003time,cadenas2010wind}. 

In this paper, we do not focus on the choices of regression models. 
Instead, based on our observation, we focus on extracting seasonality information and transforming label for better forecasting performance. 
Our proposed seasonality extraction and label distribution transform can be applied into most forecasting models, including NN and GBDT. 
\section{Conclusions}
In this paper, we studied the case of retailers' sales forecasting on Tmall---the world's leading online B2C platform. 
We first observed sales seasonality after we group different categories of retailers and Tweedie distribution after we transform the sales. 
We then designed two mechanisms, i.e., seasonality extraction and distribution transform, for sales forecasting. 
For seasonality extraction, we first adopted clustering method to group the retailers so that each group of retailers have similar features, and then  applied Fourier transform to automatically extract the seasonality for retailers in different groups. 
For distribution transform mechanism, we used Tweedie loss for regression instead of other losses that do not fit the real sale distribution. 
Our proposed two mechanisms can be used as add-ons to classic regression models, and the experimental results showed that both mechanisms can significantly improve the forecasting results.

 \bibliographystyle{splncs04}
 \bibliography{paper}

\end{document}